\documentclass{svproc}

\usepackage{url}

\usepackage[ruled,vlined]{algorithm2e}
\usepackage{booktabs}
\usepackage{geometry}
\usepackage{graphicx}
\usepackage{amssymb}
\usepackage{epstopdf}
\usepackage{textcomp}
\usepackage{xcolor}
\usepackage{subcaption}
\newcommand{\assert}[1]{\vspace{2mm}\noindent\textbf{#1}\vspace{2mm}}

\title{Can robotics research learn from computer vision research?}

\begin{document}
\mainmatter
\title{What can robotics research learn from computer vision research?}
\titlerunning{What can robotics research learn from computer vision research?} 
\author{Peter Corke \and Feras Dayoub \and David Hall \and John Skinner \and  Niko S\"{u}nderhauf}
\authorrunning{Corke, Dayoub,  Hall,  Skinner, S\"{u}nderhauf} 
\tocauthor{Peter Corke, Feras Dayoub, David Hall, John Skinner, Niko S\"{u}nderhauf}
\institute{Australian Centre for Robotic Vision, Queensland University of Technology, Australia,
\texttt{http://roboticvision.org}}

\maketitle              

\begin{abstract}

\end{abstract}

\section{A short history of computer vision and robotics research}

The fields of computer vision and robotics are both children of the artificial intelligence program that was spawned by the Dartmouth
Conference in 1956.  The first AI laboratories undertook research in both fields as well as in the application of computer vision to robotics -- then called machine vision but  nowadays called robotic vision.  Vision held the promise of solving important problems in robotics such as understanding the world, which is the input to a planning and motion control pipeline.  However, in those early days vision proved problematic: cameras were large, expensive and required special hardware in order to interface with a computer; computers were insufficiently powerful;  and obtaining good quality 3D or metric information was challenging.  Stereo vision was known and explored but in that era processing two images rather than one just made the computational problem even worse, and the quality of the results were often underwhelming. 

In the 1980s LIDARs such as the ERIM scanner became available but were large, expensive and rare. Various laboratories, notably CMU, built their own LIDARs but this was a complex route to pursue.
In the early 1990s roboticists discovered that an industrial safety barrier sensor from SICK Electro-optic could be used as a LIDAR.  It was relatively compact and gave metric data for a cross-section of the world at distances upto 50m with a resolution of the order of centimetres.  Roboticists put these sensors on various spinning and nodding mechanisms in order to create 3D point clouds.  At the time, when real-time vision processing was still not possible on regular PCs (the MMX MIMD instruction set extensions for the x86 architecture didn't arrive until the late 1990s) here was a  sensor that gave fast metric information -- it was a game changer.  

From that moment robotics and computer vision took different paths. In retrospect, many of the challenges faced by early stereo vision researchers could have been overcome at the time.  Early work was performed indoors where textureless surfaces are the norm and led to disappointing results, whereas operation outdoors would have been much more successful.  Structured lighting was known, but the modern speckle illuminator for artificially texturing indoor scenes (as used in the Kinect and RealSense sensors) was not not yet widely known.

Today, some 20 years after the great schism we see renewed interest in robotic vision.  
The computer vision and robotics research communities are each strong.  However progress in computer vision has become turbo-charged in recent years due to big data, GPU computing, novel learning algorithms and a very effective research methodology.  By comparison, progress in robotics seems slower.  It is true that robotics came later to exploring the potential of learning  -- the advantages over the well established body of knowledge in dynamics, kinematics, planning and control is still being debated, although reinforcement learning seems to offer real potential.  

However the rapid development of computer vision compared to robotics cannot be only attributed to the
former's adoption of deep learning.  In this paper we argue that the gains in computer vision is due to research methodology -- evaluation under strict constraints versus experiments; bold numbers versus videos.

\begin{figure}[t]
    \centering
    \includegraphics[width=0.6\textwidth]{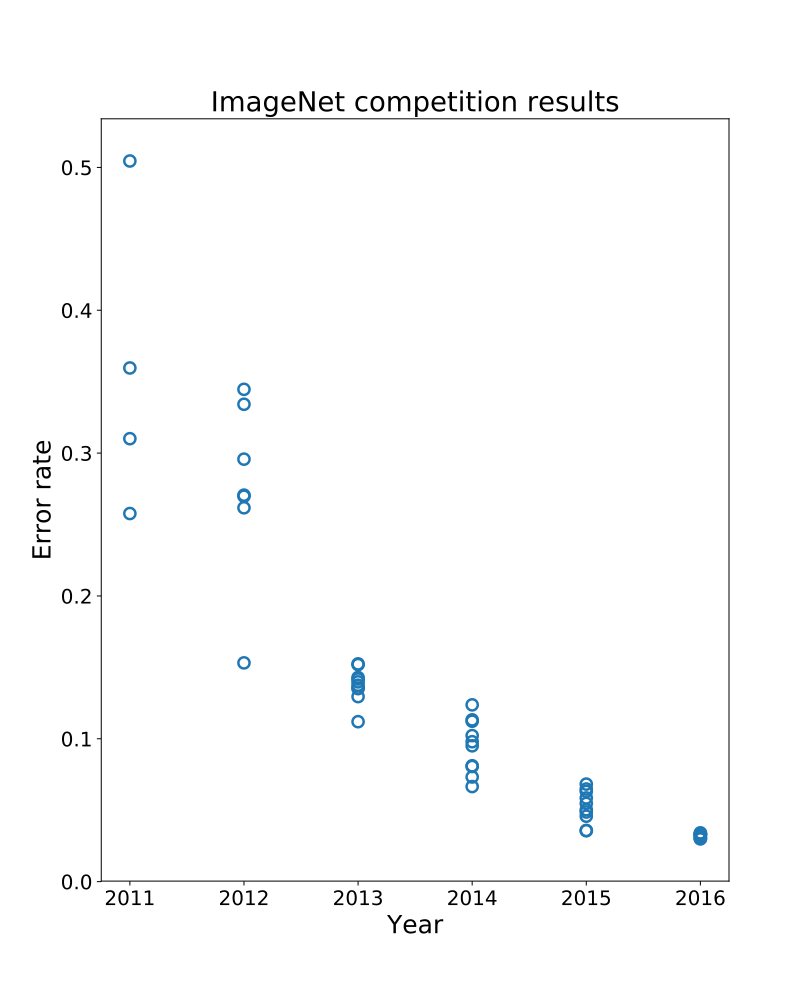}\
   \caption{Performance in ILSVRC over 2011-2016\cite{ImageNetResults}. The lowest datapoint for 2012 is AlexNet\cite{NIPS2012_4824}.}
    \label{fig:imagenet-results}
\end{figure}

\section{Recent progress in computer vision}
Progress in computer vision is more than just myriad arXiv papers with monotonically improving bold numbers.  We have seen the transition to implementations that are ubiquitous and touch vast numbers of people. For example,  face detection in every phone, image tagging in Google Photos, automated
image captioning, biometrics, medical image understanding, sports analysis and so on.

Recent progress in the area of object recognition, which is important for robotics, can be traced back to several key developments.  The first was the creation of a massive dataset called ImageNet. Today, ImageNet comprises over 14M images   with some form of annotation (what is in the image, or a bounding box of objects within the image) representing over 20,000 object categories. It was conceptualised in 2006 and was perhaps inspired by WordNet\footnote{A lexical database of English words started in 1985 to support text analysis and NLP.} and the earlier PASCAL VOC challenge based on a dataset of real images (20,000 images and 20 classes) with human-created ground truth. A key innovation of ImageNet, to avoid copyright problems was a mechanism to allow researchers to easily download the images from their original source.
This rich dataset set the scene for the second key development, a popular object recognition competition called the ImageNet Large Scale Visual Recognition Challenge (ILSVRC) which has run every year since 2010.
Performance over some of that period is shown in Figure \ref{fig:imagenet-results}.
The third key development, in 2012, was AlexNet\cite{NIPS2012_4824} an 8-layer network (5 convolutional) that gave an unprecedented improvement on the state-of-the-art and led to a huge and enduring interest in deep convolutional neural networks for problems of this kind.

Abstracting from these three developments we believe there are four important drivers which contributed to this rapid and extraordinary performance improvement:
\begin{enumerate}
\item Standard performance measures, such as top-5 or mAP.
\item Competitions, such as ILSVRC.
\item Zero-delay and guaranteed dissemination, through arXiv, has increased the rate of innovation.  Traditional conference and journal outlets exhibit a long latency between results and publication, and low-acceptance rates and noise in the review process\cite{doi:10.1177/014107680609900414} reduce the chances that important results are shared.
\item Wealth creation, these research results can already be monetized for face detection in phone cameras, online photo album searching and tagging, biometrics, social media and advertising.
\end{enumerate}

\assert{Assertion 1: standard datasets + competition + rapid dissemination \textrightarrow rapid progress}

The robotics community has also pursued the creation of datasets which can be used as training data for perception systems, or as   benchmarks in other papers. Some notable examples include the Oxford Robotcar dataset~\cite{RobotCarDatasetIJRR} of self-driving car sensor data and the Yale-CMU-Berkeley Dataset for robotic manipulation research~\cite{7251504}. 
However the bulk of robotics datasets don't achieve much impact, perhaps because there are just too many to choose from so that none gets ahead, perhaps because of roboticist's penchant for reinventing wheels, perhaps a mistaken belief that including a dataset gives a paper an advantage at review time, overestimating the generality and utility of the dataset, and a lack of effort on documentation and tools to make the dataset truly usable.

\assert{Assertion 2: datasets without competitions will have minimal impact on progress}

\section{Applying the computer vision magic to robotics}
As roboticists, we might ask: how can we get into the fast lane of innovation where we see our computer vision colleagues?  At the lowest level we can just ``take'' the deep-learnt perception stack from the vision community and use it on our robot, and continue doing robotics as we currently do.  A bolder approach would to take the whole methodology of standard datasets, standard metrics, competitions and rapid dissemination.

\begin{figure}
    \centering
    \includegraphics[width=0.8\linewidth]{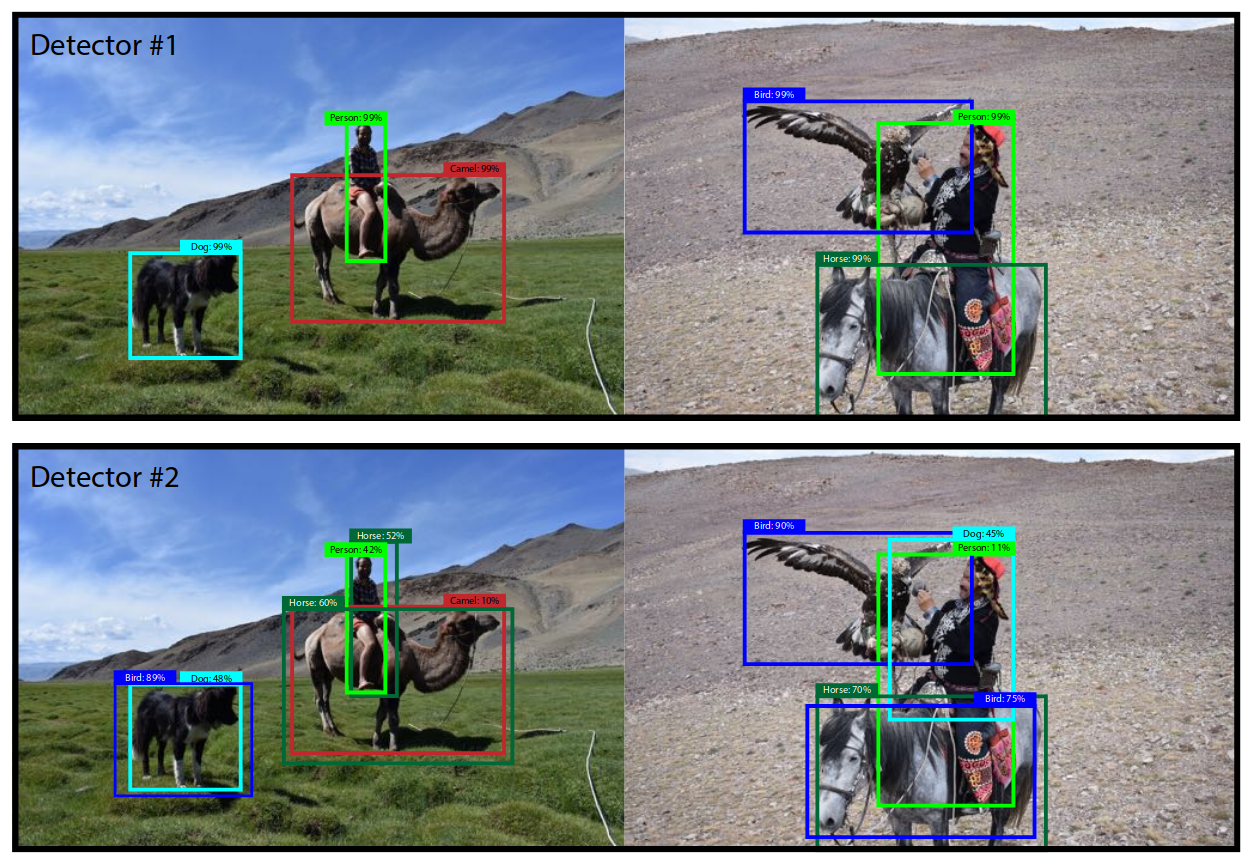}
    \caption{Output of two hypothetical detectors that have qualitatively different outputs but under mAP are both considered perfect and equal. Reproduction of figure from~\cite{redmon2018yolov3}.}
    \label{fig:yolo_map}
\end{figure}

\begin{figure*}[t]
    \centering
    \begin{subfigure}[t]{\linewidth}
    \centering
        \begin{subfigure}[b]{0.2\linewidth}
            \includegraphics[width=\textwidth]{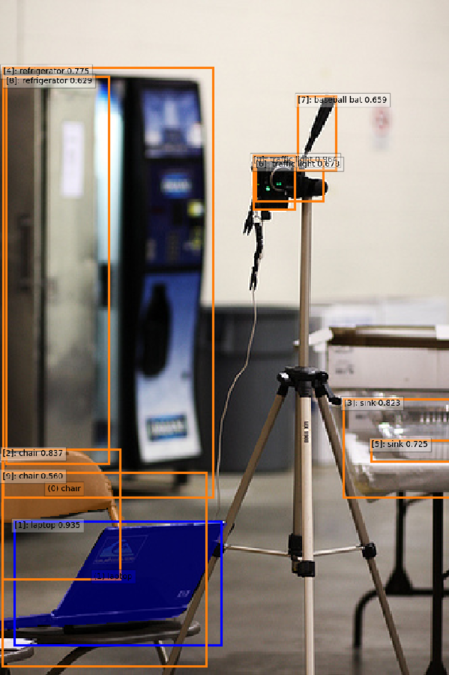}
        \end{subfigure}
        \begin{subfigure}[b]{0.2\linewidth}
            \includegraphics[width=\textwidth]{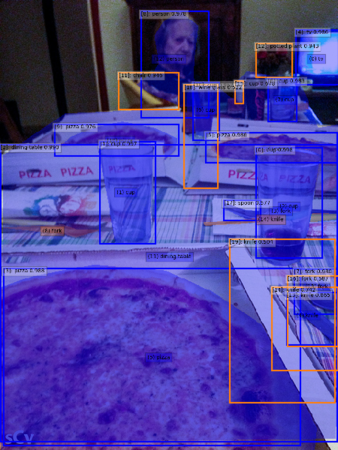}
        \end{subfigure}
        \begin{subfigure}[b]{0.2\linewidth}
            \includegraphics[width=\textwidth]{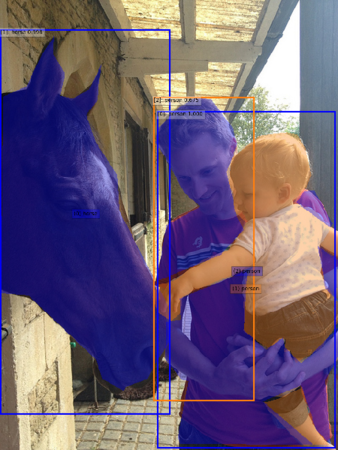}
        \end{subfigure}
        \begin{subfigure}[b]{0.2\linewidth}
            \includegraphics[width=\textwidth]{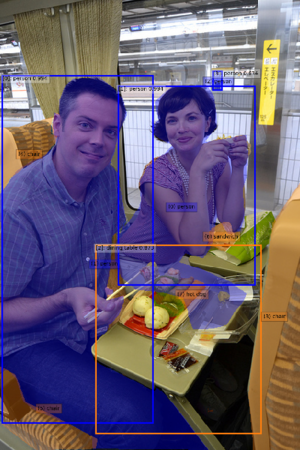}
        \end{subfigure}
    \caption{mAP score 37.4\%}
    \end{subfigure}
    \begin{subfigure}[t]{\linewidth}
    \centering
        \begin{subfigure}[b]{0.2\linewidth}
            \includegraphics[width=\textwidth]{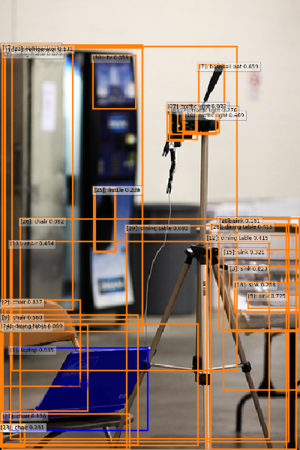}
        \end{subfigure}
        \begin{subfigure}[b]{0.2\linewidth}
            \includegraphics[width=\textwidth]{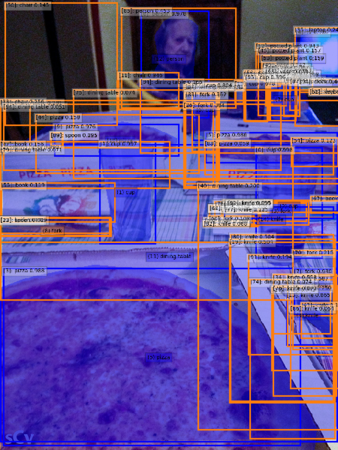}
        \end{subfigure}
        \begin{subfigure}[b]{0.2\linewidth}
            \includegraphics[width=\textwidth]{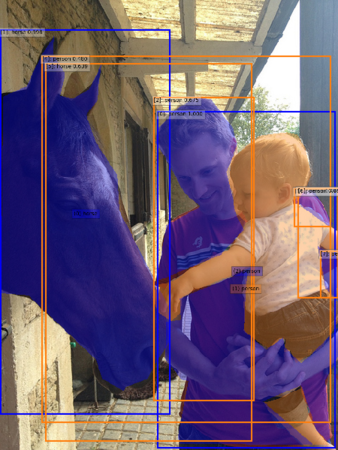}
        \end{subfigure}
        \begin{subfigure}[b]{0.2\linewidth}
            \includegraphics[width=\textwidth]{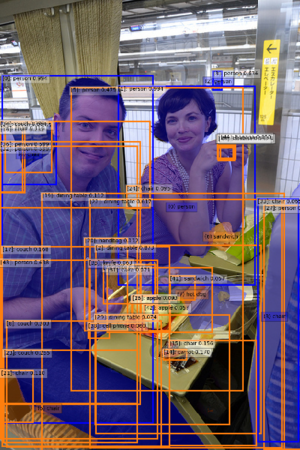}
        \end{subfigure}
    \caption{mAP score 39.0\%}
    \end{subfigure}
    \caption{Detection output from two detectors with mAP scores 37.4\% (a) and 39.0\% (b) respectively. Visualizations of true and false positives (blue and orange boxes) and true and false negatives (blue and orange masks) are based on probability-based detection quality. Visualization shows large false number of false positives associated with detector with higher mAP score. Reproduction of figure from supplementary material of~\cite{hall2020pdq}}
    \label{fig:pdq_fp}
\end{figure*}

\pagebreak As roboticists we would likely have some objections to that methodology, in particular the dataset and metric aspects:
\begin{itemize}
\item Does ImageNet performance actually translate to the real world? 
A growing body of research\cite{recht2019imagenet,torralba2011unbiased,barbu2019objectnet} is showing that the state-of-the-art deep neural networks do not generalise well and incur a drop in their accuracy in the range of 10\% to 14\% when tested on data outside of their benchmark datasets.
\item Is top 5 or mAP really a good measure for robot performance? 
 Top 5 classification accuracy suits human assistance applications with no localization components like leaf recognition~\cite{kumar2012}. 
 This does not apply to robotics which should operate with minimal human interaction.
 While the mAP measure considers both classification and localization, it is increasingly shown to have some non-intuitive behaviours.
 For example \cite{redmon2018yolov3} shows that two detection outputs can qualitatively have very different performance yet both attain perfect mAP scores, see Figure \ref{fig:yolo_map}. 
 This is reinforced in~\cite{hall2020pdq} where it was shown that detector outputs with high-volume, (albeit lower-confidence) false positives could outperform detectors with comparitively lower false positive counts, see Figure~\ref{fig:pdq_fp}.

\item How do we estimate uncertainty? As robots operate in the physical world, it can be unsafe to operate with absolute confidence both semantically and spatially. Both spatial and semantic uncertainty should be meaningfully estimated and evaluated.

\item How do we deal with dataset bias?  The images in ImageNet, Coco and many other common datasets are mostly taken by humans and have objects centred in the frame, unoccluded and with good exposure. There is also a selection bias in any scene which means that the distribution of possible views in a given environment is far from uniform. Recent results~\cite{barbu2019objectnet} suggest that typical classifiers experience a 40\% drop in performance when presented with objects from unusual viewpoints.

\item A common assumption in deep learning is that trained models will be deployed under
\emph{closed-set} conditions, i.e. the classes encountered during deployment are known and exactly the same as during training. However, robots often have to operate in ever-changing, uncontrolled real-world environments, and will inevitably encounter instances of classes, scenarios, textures, or environmental conditions that were not covered by the training data. This regime is referred to as \emph{open-set} conditions~\cite{torralba2011unbiased,bendale2015towards} and often requires approaches that can identify unknown inputs, e.g. by utilizing epistemic uncertainty~\cite{blum2019fishyscapes,kendall2017uncertainties}.

\item Can a dataset-driven methodology be applied to an embodied agent?  In ILSVRC the images that are classified are independent of one another, but a robot is a closed-loop system.  Each image it perceives has overlap with the previous image, but more importantly any image captured is a function of all the previous control actions and thus all preceeding images.
\end{itemize}

\pagebreak 
\noindent Computer vision research is driven by a methodology of \emph{evaluation}:
\begin{quote}
\textit{the systematic assessment of the operation and/or the outcomes of a system, compared to a set of explicit or implicit standards, as a means of contributing to the improvement of the system} (paraphrased from \cite{Weiss98})
\end{quote}
whereas robotics research purports to be driven by \emph{experimentation} and in fact many conferences and journals
will not accept papers without experimental results.
\begin{quote}
\it  An experiment is a scientific procedure undertaken to make a discovery, test a hypothesis, or demonstrate a known fact.\\[10pt]

\it  A hypothesis:
\begin{enumerate}
\item is a tentative, testable answer to a scientific question. 
\item leads to one or more predictions that can be tested
\end{enumerate}
\end{quote}
However it is questionable whether roboticists are really conducting experiments because the hypothesis in most published robotics experiments is rarely stated.
With \emph{evaluation} we can say something profound like ``\textit{my algorithm works x\% better than another algorithm with respect to performance measure y}'' whereas with an experiment we can say ``\textit{my robot worked}" but left unsaid is how many times it worked, the reliability and
conditions under which it worked.

\assert{Assertion 3: to drive progress in robotics we should change our mindset from experiment to evaluation}

\section{Applying the evaluation methodology to robotics research}
In order to evaluate the performance of a robot we need to agree upon a robotic task and relevant performance metrics. 
Standard tasks conducted in standard environments can help to put results on an equal footing. 
Some robot competitions provide standard tasks and performance measures, for example Learning Applied to Ground Robots (LAGR), Robocup, Amazon Robotics Challenge, DARPA Robotics Challenge, and DARPA Urban challenge.

Performance will also depend on the robot itself: the quality of its sensors, the capability of its actuators, the sample rate
achievable by its processing system.
Standard platforms can help to factor out the robot variable.
For robot arms, the standard was once the Unimation Puma 560 and today is perhaps a Universal robot or a Franka-Emika Panda. 
For mobile robots, there have been standard platforms such as the LAGR robot, Robocup standard league, and iCub to name just a few.
The Robotarium\cite{robotarium} provides standard robots and environment but the task can be chosen by the user.

However we cannot escape the reality that the performance of a sensor-based robot is stochastic.
Each run of the robot is unrepeatable since 
no two images of the same scene will be equal due to photo well shot and thermal noise, lighting flicker,
other sensor noise, and the 
initial pose of the robot cannot be set to achieve precise image alignment with earlier runs.
Consider the simple example of a vision-based robot: variation in the first image (due to shot noise, lighting variation, initial camera pose) leads to different control outcomes.  The second image, taken from the resulting different poses will be further different and from there the performances will diverge.

\pagebreak Real-world robot performance measures are ultimately non-repeatable.  The only way to achieve repeatability in evaluation is to
adopt simulation which allows:
\begin{itemize}
\item the comparison of different algorithms on the same robot, environment \& task
\item estimating the distribution in algorithm performance due to sensor noise, initial condition, etc.
\item investigating the robustness of algorithm performance due to environmental factors
\item regression testing of code after alterations or retraining
\end{itemize}

\assert{Assertion 4: simulation is the only way in which we can repeatably evaluate robot performance}

The next two sections provide some examples of simulation for evaluating the performance of robotic algorithms.

\subsection{Evaluating the performance of ORB-SLAM}\label{sec:orbslam}

\begin{figure}[t]
    \centering
    \includegraphics[width=0.8\textwidth]{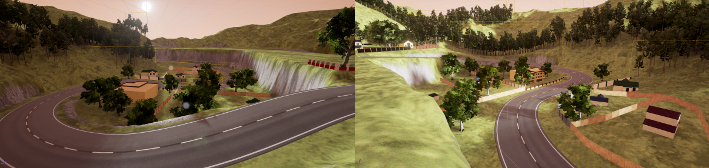}
   \caption{Simulated environment used to evaluate ORB-SLAM performance.}
    \label{fig:unreal-world}
\end{figure}
\begin{figure}[t]
    \centering
    \begin{tabular}{l}
    \includegraphics[width=0.8\textwidth]{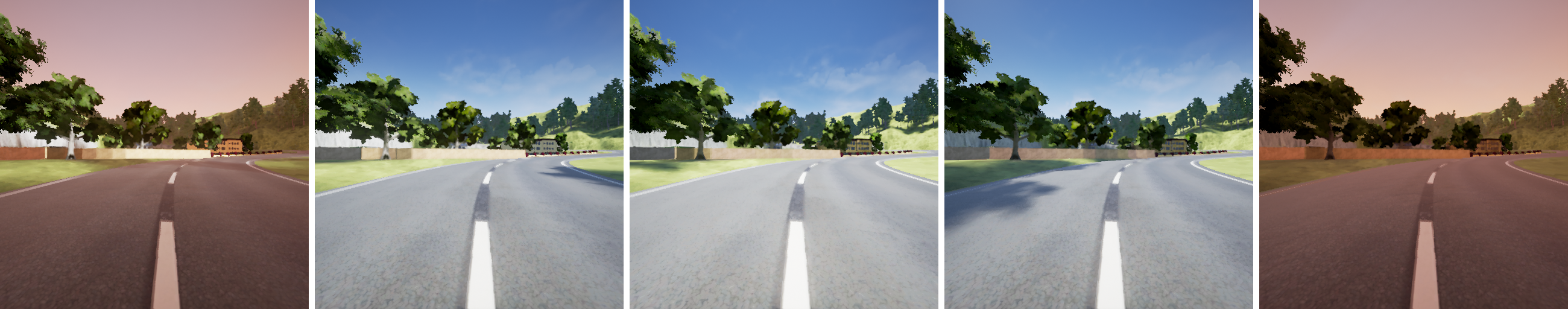}\\
    \includegraphics[width=0.8\textwidth]{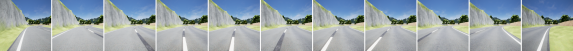} \\
    \includegraphics[width=0.8\textwidth]{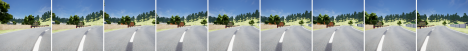} \\
    \includegraphics[width=0.8\textwidth]{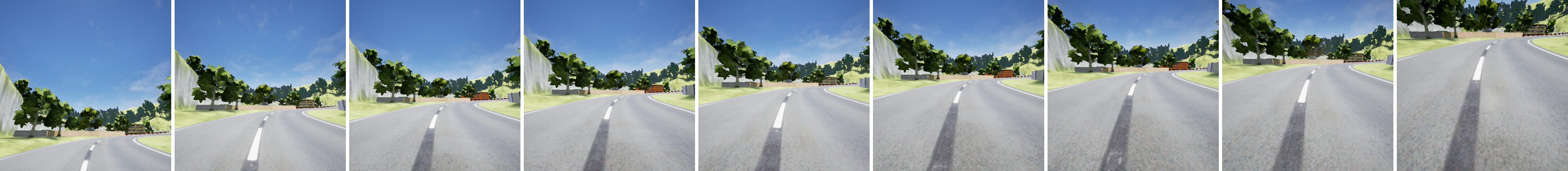} 
    \end{tabular}
   \caption{Example images at the same point showing, from top to bottom, the effect of times of day, lateral offset, camera yaw angle and camera pitch angle.}
    \label{fig:images-stacked}
\end{figure}

In \cite{skinner} we evaluated the localization performance of ORB-SLAM against variation in viewpoint and lighting level.
ORB-SLAM \cite{Montiel2015} is a feature-based monocular SLAM system, that performs feature-based visual feature tracking, place recognition, mapping and loop closure using ORB features.
The images were generated using Unreal Engine 4 by Epic Games\footnote{\url{https://www.unrealengine.com}} which is a state-of-the-art simulation tool developed for gaming which  allows the creation of complex 3-dimensional worlds that are realistically rendered, see Figure \ref{fig:unreal-world}.  The  view from a camera at any arbitrary pose can be obtained, enabling us to mimic the motion of a robot through the environment, and the robot can have one or many cameras. We can also change the illumination conditions, adjusting the position of the sun, the cloud conditions, artificial light sources and atmospheric conditions such as fog. Because exact ground truth is known, camera poses and object positions estimated by robotic vision algorithms can be compared against the simulation state information.

\begin{figure}[t]
\centering
\includegraphics[width=.3\textwidth]{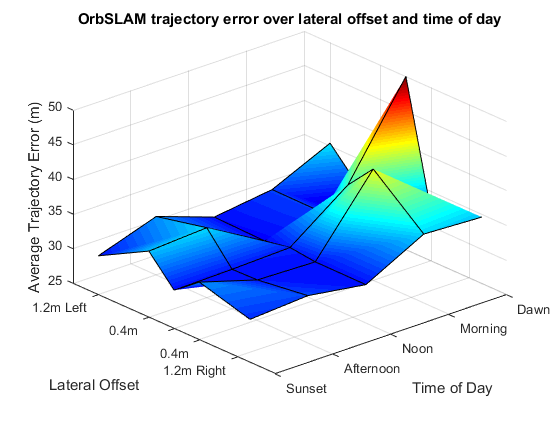}\hfill
\includegraphics[width=.3\textwidth]{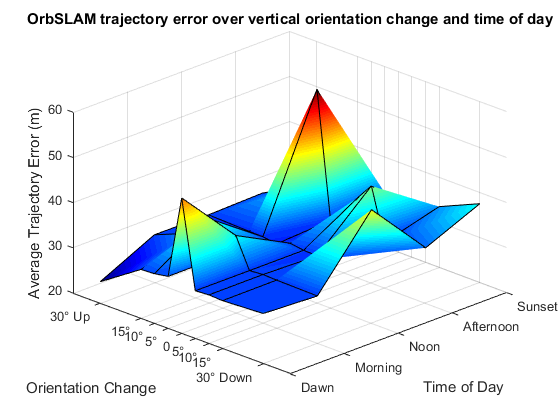}\hfill
\includegraphics[width=.3\textwidth]{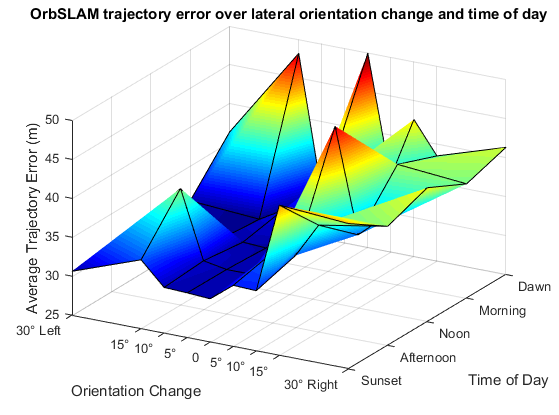}
\caption{OrbSLAM average trajectory error over both time of day and lateral offset and camera orientation change}
\label{fig:orbslam-results}
\end{figure}

To investigate the viewpoint invariance and time of day invariance of ORB-SLAM we created many variations of the same camera trajectory, at different times of day, with different lateral offsets, and with different camera pitch and yaw angles, see Figure \ref{fig:images-stacked}.
For each combination of condition we compared the ORB-SLAM estimated trajectory with the ground truth to calculate the Absolute Trajectory Error after aligning their scale and the results are summarised in Figure \ref{fig:orbslam-results}.
We can see that the behaviour is not perfect, and that the conditions for poor performance are complex and non-obvious.
This characteristic would be hard to discover any other way than simulation.

\subsection{A simulation-based robotic vision competititon}\label{sec:rvchallenge}

In 2018, we developed and released a new ongoing Robotic Vision Challenge~\footnote{\url{https://www.roboticvisionchallenge.org}} that focuses on object detection, but addresses robotic vision-specific problems that 
are not well covered by current object detection challenges in computer vision challenges such as COCO\cite{lin2014microsoft}, or ILSVRC\cite{ilsvrc:russakovsky2014imagenet}. Specifically, there are three major differences between our new challenge differs and the established computer vision object detection challenges:
\begin{enumerate}
    \item We explicitly evaluate spatial and semantic \emph{uncertainty} through a novel evaluation measure coined Probabilistic Detection Quality (PDQ)~\cite{hall2020pdq}. This forces participants to devise new \emph{probabilistic} approaches to object detection that can provide uncertainty information.
    \item Our dataset consists of video sequences captured by simulated robots in a variety of semantically rich indoor environments. Object detection methods that understand that the camera is embodied in a robot can exploit temporal information and coherence between consecutive frames, and thus achieve higher scores.
    \item We focus on object detection in realistic open-set conditions (i.e. during test time, object classes that were never seen during training can appear in the field of view of the robot). PDQ penalises false positive detections on such open-set classes, thus motivating participants to develop detectors that are robust in open-set conditions.
\end{enumerate}

We identified these three major research challenges -- the need for uncertainty, evaluation in open-set conditions, and (temporal) embodiment -- in earlier work, where we discussed the limits and potentials of deep learning in robotics~\cite{sunderhauf2018limits}. We used these as the guiding principles when devising the proposed Robotic Vision Challenge.

Failures or mistakes in object detection can lead to catastrophic outcomes that not only risk the success of the robot's mission, but potentially endanger human lives. We can distinguish four major types of object detection failures: 
\begin{enumerate}
    \item failing to detect an object, 
    \item assigning a wrong class label to a detected object, 
    \item badly localising an object, e.g. only detecting parts of it, and 
    \item erroneously detecting non-existing objects. 
\end{enumerate}
Despite a lot of progress over recent years, today's state-of-the-art object detectors are still prone to all of these failure types. A important factor in this is that existing detectors are not built to express \emph{uncertainty}, which often leads to \emph{overconfident} detections. While probabilistic techniques that can properly incorporate uncertainty have been established in robotics for many years~\cite{thrun2005probabilistic}, such methods are rare in computer vision, and have been largely ignored by the big computer vision challenges. 
A probabilistic approach for evaluation of object detection would adhere to the following principles: 

\begin{itemize}
\item it is ok to be wrong, as long as uncertainty is high,
\item being correct is worth less when uncertainty is high,
\item spatial and label uncertainty both contribute to the overall certainty of a detection.
\end{itemize}

\begin{figure}[t]
    \centering
    \includegraphics[width=0.7\textwidth]{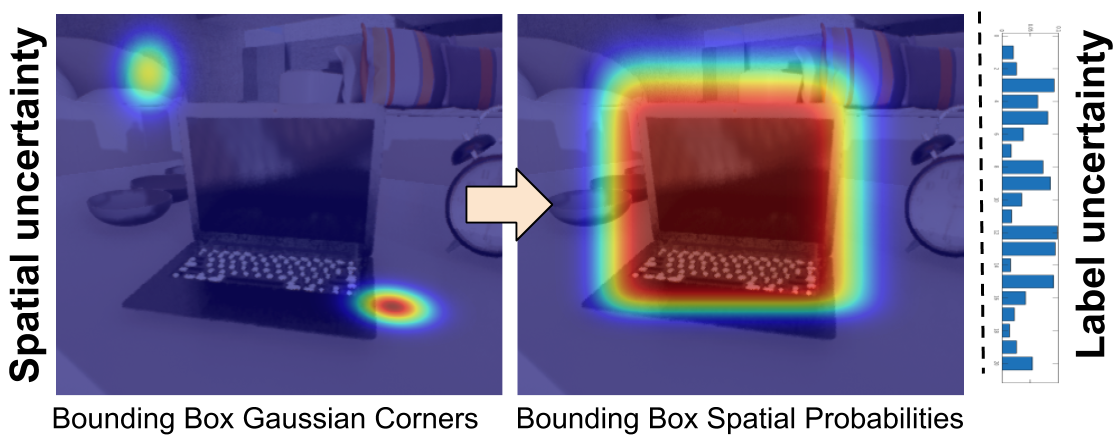}
  \caption{Probabilistic object detectors express semantic and spatial uncertainty.
Object locations are reported as probabilistic bounding boxes where corners are modelled as 2D Gaussians (left) used to express a spatial uncertainty over the pixels (centre). Semantic uncertainty is represented as a full label probability distribution (right).}
    \label{fig:pdq-1}
\end{figure}

In order to foster progress towards a probabilistic approach to object detection, our Robotic Vision challenge introduces the new task of \emph{Probabilistic Object Detection}~\cite{hall2020pdq}. The aim is to detect objects in images while accurately quantifying the spatial and semantic uncertainties of the detections. Probabilistic Object Detection poses key challenges that go beyond the established conventional object detection.
Firstly, the detector must quantify its \emph{spatial uncertainty} by reporting \emph{probabilistic} bounding boxes where the box corner positions are normally distributed as shown in Figure~\ref{fig:pdq-1} -- this induces a spatial probability distribution over the image for each detection.
Secondly, the detector must  reliably quantify its \emph{semantic uncertainty} by providing a full probability distribution over the known classes for each detection. 

\begin{table}[b]
\centering
\caption{Leaderboard of the top four participating teams, as evaluated by the PDQ measure~\cite{hall2020pdq}}
\label{tab:pdq-leaderboard}
\begin{tabular}{@{}r|cllllll@{}}
\toprule
Place & \multicolumn{1}{c|}{PDQ Score} & \multicolumn{1}{l|}{\begin{tabular}[c]{@{}l@{}}Average\\ Overall\\ Quality\end{tabular}} & \multicolumn{1}{l|}{\begin{tabular}[c]{@{}l@{}}Average \\ Spatial\\ Quality\end{tabular}} & \multicolumn{1}{l|}{\begin{tabular}[c]{@{}l@{}}Average\\ Label\\ Quality\end{tabular}} & \multicolumn{1}{l|}{\begin{tabular}[c]{@{}l@{}}True \\ Positives\end{tabular}} & \multicolumn{1}{l|}{\begin{tabular}[c]{@{}l@{}}False \\ Positives\end{tabular}} & \begin{tabular}[c]{@{}l@{}}False \\ Negatives\end{tabular} \\ \midrule
1st \cite{wangaugpod} & 22.56 & 0.60 & 0.45 & 1.00 & 152967 & 113620 & 143400 \\
2nd \cite{ammirato2019mask} & 21.43 & 0.66 & 0.50 & 0.99 & 125332 & 90853 & 171035 \\
3rd \cite{liteamgl} & 20.01 & 0.65 & 0.49 & 1.00 & 118678 & 91735 & 177689 \\
4th \cite{MorrisonPDQ} & 14.65 & 0.57 & 0.47 & 0.85 & 87438 & 48327 & 208929 \\ \bottomrule
\end{tabular}
\end{table}

To evaluate how well detectors perform on this challenging task, we introduce a new evaluation measure: Probability-based Detection Quality (PDQ)~\cite{hall2020pdq}. PDQ explicitly evaluates the reported probability of the true class via its \emph{Label Quality} component. 
Furthermore, PDQ contains a \emph{Spatial Quality} term that evaluates how well a detection's spatial probability distribution matches the true object. 

The Robotic Vision Challenge uses video sequences generated from simulation using Unreal Engine 4 with a modified version of NVidia's Dataset Synthesizer\footnote{\url{https://github.com/jskinn/Dataset_Synthesizer}}, with sequences spanning multiple physical environments, day- and night-time lighting conditions, and different camera heights.
These video sequences are divided into a \textbf{test} set, a \textbf{test\_dev} set, and a \textbf{validation} set, with ground truth made publicly available for the \textbf{validation} set.
The \textbf{test} set is used for the first of our fixed-time challenges where top competitors were awarded prizes at a workshop we organised during CVPR 2019. Table~\ref{tab:pdq-leaderboard} contains information about the top four teams and their scores.

The \textbf{test\_dev} set provides an ongoing benchmark that can be used as a baseline to drive future research. No training set is provided for this challenge, and participants are encouraged to train on whatever data seems appropriate.
In doing so, we hope to avoid dataset bias and encourage competitors to develop systems that can generalise to the test data, rather than fitting to the particulars of this challenge.

\assert{Assertion 5: we can use new competitions (and new metrics) to nudge the research community}

We believe that research challenges, evaluation metrics, and competitions can make the limits of the current state-of-the-art visible, motivate the community to address these limits, and track progress over time. This has been demonstrated time and again by big computer vision challenges and competitions that had a significant influence on the advancements in computer vision in recent years. However, it is important that we as roboticists identify the specific problems that are unique to robotic vision. Where these open research problems differ from computer vision, we have to create our own evaluation environments, protocols, metrics, datasets or simulations. Relying purely on computer vision benchmarks, and lacking meaningful standardised evaluation protocols and benchmarks for these research challenges is a significant roadblock for the evolution of robotic vision, and impedes reproducible and comparable research in our community.

\subsection{Considerations for the future of evaluation in robotics}
Moving forward, it is important to acknowledge the potential hazards in robotics research which which could arise through challenges or competitions.

\vspace{3mm}
\noindent \textbf{Choice of challenge tasks} - When developing problems for standardized evaluation, it is important that the problems are relevant, difficult, and have wide-ranging implications.
Challenges should focus on problems that can be generally applied to different robot configurations and settings.
It should be acknowledged that standardized challenges will not be applicable for all areas of research, particularly where solutions would focus on novel hardware design.
The robotics community should not ignore these problems in their use of benchmark challenges.

\vspace{3mm}
\noindent \textbf{Overfitting to challenges} - As has been seen in the computer vision community, it can become easy to overfit for a specific challenge such that use of a different datasets can greatly reduce performance~\cite{recht2019imagenet,torralba2011unbiased,barbu2019objectnet}.
Such a situation could also occur with robotics challenges.
Widespread implementation of any approach should be considered and commented on when using a robotics challenge for research.
Challenge results (particularly those attained through simulation) should not be viewed as absolute proof of a method's applicability to all real-world settings.
Challenges should be continuously updated to avoid stagnation in fields of research, and reduce overfitting to any individual challenge.
This could be achieved by new environmental setups, new evaluation metrics, or entirely new challenges relating to emerging aspects of a research area.
Of course this requires time and resources from members of the community but is crticial to ensuring the nature of evaluation-based robotics research continues and remains relevant.

In summary, when using standardized challenges for robotics research, it is important to avoid the mentality that the results of the challenges can be viewed entirely in isolation. Robotics, as a heavily application-focused domain of research, should always consider how methodologies will work outside the more constrained setting of a challenge for more widespread application.

\section{Revisiting Assertion 1}
Assertion 1 attributes progress in computer vision to various factors including competitions but ``competition" needs to be unpacked.  The key characteristics of a good competition are:
\begin{itemize} 
\item clear rules: the evaluation metric
\item a prize (fame or fortune): to induce rivalry
\item competitors: ideally many who are both capable and motivated
\end{itemize}
More competitors and more rivalry brings a diversity of approaches and progress with respect to the competition rules.

\assert{Assertion 1\textquotesingle: standard datasets + competition (evaluation metric + many smart competitors + rivalry) + rapid dissemination \textrightarrow ~rapid progress}

An unfortunate reality is that the number of people who can play in robotics is orders of magnitude smaller than it is for computer vision.
The requirements for computer vision is simply smart graduate students, some open-source software tools, access to GPUs and the ability to write papers quickly.
Robotics, by contrast, requires all those things plus robots, sensors, space to operate safely,
the smart person in the lab who knows how to run the robot, mastery of ROS, lots of time and plenty more.

\begin{figure}[t]
    \centering
    \includegraphics[width=0.5\textwidth]{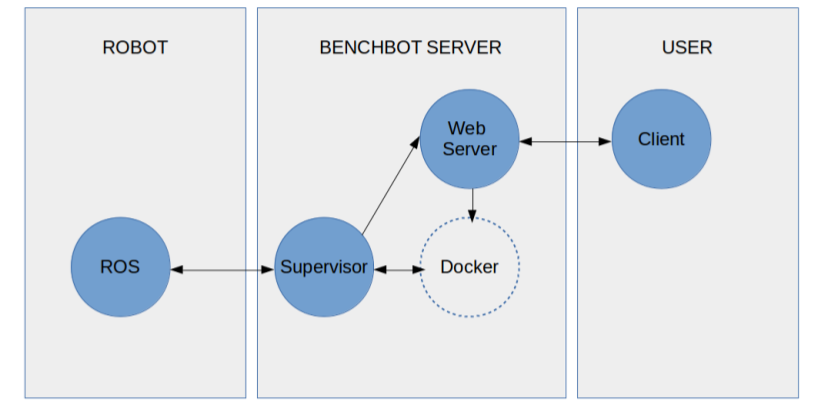}
  \caption{An overview of the BenchBot system architecture. The client uses BenchBot's API to test their algorithm on our robots and packages the code in a Docker file that is upload to our server through a website. The code is then executed on the robot and the evaluation results is returned to the client at the end of the task.}
    \label{fig:bechbot-archi-1}
\end{figure}

\subsection{Benchbot: access to robots for everyone}\label{sec:benchbot}

The aim of BenchBot is to democratise robotics by giving researchers access to real robots that  they might otherwise not be able to access. 
It eliminates the needs to: own a robot, have space and staff to operate it, and mastery of ROS.
BenchBot is an online portal that allows researchers anywhere in the world to remotely test their machine learning and computer vision systems onboard various types of real robots in real environments in a fair and equal setting.

\begin{algorithm}[b]
\SetAlgoLined
\KwResult{evaluation\_score}
benchbot          = BenchBot()\;
agent = Agent(benchbot.actions)\;
observations , env = benchbot.reset()\;
 \While{not agent.is\_done()}{
  next\_action = agent.policy\_function(observations)\;
  observations , info = env.step(action)\;
 }
 evaluation\_score = benchbot.eval(agent.get\_results())\;
 \caption{BenchBot Main Loop}
 \label{Algo:bechbot-loop}
\end{algorithm}

 Our code is open-source which allows other research labs around the world to replicate our BenchBot setup, opening more online portals to accommodate more researchers to test their algorithms remotely on real robotic hardware. The ultimate goal is that BenchBot becomes a standardized test for robotics research that allow various algorithms to be tested on real robots. 

An overview of the BenchBot architecture is shown in Figure~\ref{fig:bechbot-archi-1}. The client (i.e. the researcher who wants to test a computer vision algorithm on our robots), implements the function \texttt{policy\_function()} 
as shown in Algorithm~\ref{Algo:bechbot-loop}. This code is then packaged as a Docker file and uploaded  to our server through a website. The client's code is then executed on the robot by the supervisor, and the loop in Algorithm 1 is run until the robot achieves its task or fails.

\begin{figure}[tp]
    \centering
    \includegraphics[width=0.9\textwidth]{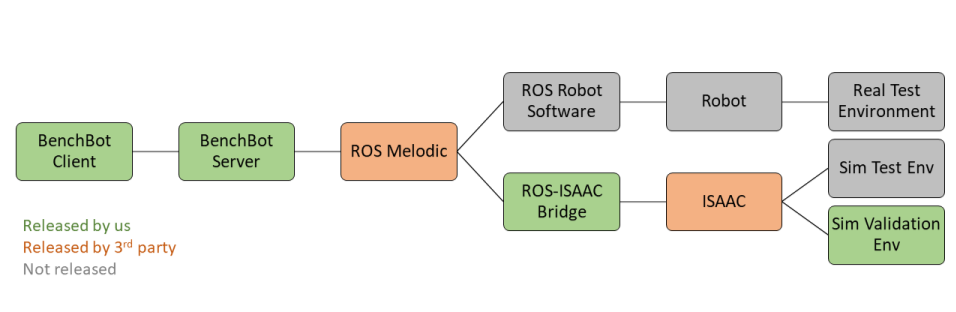}
  \caption{BenchBot can sit on top of ROS on real robots or robotic simulators that use ROS such as NVIDIA Isaac enabling the same client code to run seamlessly on real or simulated robots.}
    \label{fig:bechbot-archi-2}
\end{figure}

Our BenchBot API can sit on top of robotic simulators that  use ROS such as NVIDIA Isaac. From the user point of view, the code to operate a real or simulated robot is the same, as shown in Figure~\ref{fig:bechbot-archi-2}.

\section{Conclusions}
The computer vision and robotics research communities had a shared beginning but have diverged in terms of conferences and journals, research methodology and research rate.
From a robotics perspective it seems that computer vision is in the fast lane while we are stuck in the slow lane.
Roboticists hold a fundamental belief in the importance of experimentation but could it be that experiments are actually holding us back? Or is it that we are doing experiments poorly?

Computer vision research is based on evaluation rather than experimentation. Many aspects of their methodology could be applied to robotics but in order to achieve meaningful comparison and repeatability we need to adopt simulation (the robotics equivalent of a standard image dataset).  We have shown how simulation can be used to rigorously evaluate a standard SLAM algorithm under varying lighting and viewpoint conditions (Section \ref{sec:orbslam}) and to
enable a new competition to advance the performance of robotic vision systems with respect to dealing with uncertainty (Section \ref{sec:rvchallenge}).
We recognise the importance of evaluation on real robots but also the many challenges that this involves, and to increase the accessibility of real robots to the vision research community we have developed the BenchBot system (Section \ref{sec:benchbot}).

This paper has made a number of (bold) assertions, summarised below, which should be debated within our community to enhance the rate at which we collectively innovate:
\begin{enumerate}
    \item standard datasets + competition (evaluation metric + many smart competitors + rivalry) + rapid dissemination \textrightarrow ~rapid progress

    \item datasets without competitions will have minimal impact on progress
    \item to drive progress we should change our mindset from experiment to evaluation
    \item simulation  is  the  only  way  in  which  we  can  repeatably  evaluate  robot performance
    \item  we can use new competitions (and new metrics) to nudge the research community
\end{enumerate}

\section*{Acknowledgements}

We thank the organizers of ISRR2019, in Hanoi, for the invitation to present a first pass of these ideas in a Distinguished Talk.
This research was conducted by the Australian Research Council Centre of Excellence for Robotic Vision (project number CE140100016).

\bibliography{isrr2019}

\begin{thebibliography}{10}
\providecommand{\url}[1]{\texttt{#1}}
\providecommand{\urlprefix}{URL }

\bibitem{ImageNetResults}
Performance in {ILSVRC} over 2011 to 2016,
  \url{https://commons.wikimedia.org/wiki/File:ImageNet_error_rate_history_(just_systems).svg}

\bibitem{NIPS2012_4824}
Krizhevsky, A., Sutskever, I., Hinton, G.E.: Imagenet classification with deep
  convolutional neural networks. In: Pereira, F., Burges, C.J.C., Bottou, L.,
  Weinberger, K.Q. (eds.) Advances in Neural Information Processing Systems 25,
  pp. 1097--1105. Curran Associates, Inc. (2012),
  \url{http://papers.nips.cc/paper/4824-imagenet-classification-with-deep-convolutional-neural-networks.pdf}

\bibitem{doi:10.1177/014107680609900414}
Smith, R.: Peer review: A flawed process at the heart of science and journals.
  Journal of the Royal Society of Medicine  99(4),  178--182 (2006),
  \url{https://doi.org/10.1177/014107680609900414}, pMID: 16574968

\bibitem{RobotCarDatasetIJRR}
Maddern, W., Pascoe, G., Linegar, C., Newman, P.: {1 Year, 1000km: The Oxford
  RobotCar Dataset}. The International Journal of Robotics Research (IJRR)
  36(1),  3--15 (2017), \url{http://dx.doi.org/10.1177/0278364916679498}

\bibitem{7251504}
{Calli}, B., {Singh}, A., {Walsman}, A., {Srinivasa}, S., {Abbeel}, P.,
  {Dollar}, A.M.: The {YCB} object and model set: Towards common benchmarks for
  manipulation research. In: 2015 International Conference on Advanced Robotics
  (ICAR). pp. 510--517 (July 2015)

\bibitem{redmon2018yolov3}
Redmon, J., Farhadi, A.: Yolov3: An incremental improvement. arXiv preprint
  arXiv:1804.02767  (2018)

\bibitem{hall2020pdq}
Hall, D., Dayoub, F., Skinner, J., Zhang, H., Miller, D., Corke, P., Carneiro,
  G., Angelova, A., Sünderhauf, N.: Probabilistic object detection: Definition
  and evaluation. In: 2020 IEEE Winter Conference on Applications of Computer
  Vision. IEEE (2020)

\bibitem{recht2019imagenet}
Recht, B., Roelofs, R., Schmidt, L., Shankar, V.: Do {ImageNet} classifiers
  generalize to {ImageNet}? arXiv preprint arXiv:1902.10811  (2019)

\bibitem{torralba2011unbiased}
Torralba, A., Efros, A.A., et~al.: Unbiased look at dataset bias. In: CVPR.
  vol.~1, p.~7. Citeseer (2011)

\bibitem{barbu2019objectnet}
Barbu, A., Mayo, D., Alverio, J., Luo, W., Wang, C., Gutfreund, D., Tenenbaum,
  J., Katz, B.: Objectnet: A large-scale bias-controlled dataset for pushing
  the limits of object recognition models. In: Advances in Neural Information
  Processing Systems. pp. 9448--9458 (2019)

\bibitem{kumar2012}
Kumar, N., Belhumeur, P.N., Biswas, A., Jacobs, D.W., Kress, W.J., Lopez, I.C.,
  Soares, J.V.: Leafsnap: {A} computer vision system for automatic plant
  species identification. In: Computer {Vision}–{ECCV} 2012, pp. 502--516.
  Springer (2012), 00067

\bibitem{bendale2015towards}
Bendale, A., Boult, T.: Towards open world recognition. In: Proceedings of the
  IEEE Conference on Computer Vision and Pattern Recognition. pp. 1893--1902
  (2015)

\bibitem{blum2019fishyscapes}
Blum, H., Sarlin, P.E., Nieto, J., Siegwart, R., Cadena, C.: The fishyscapes
  benchmark: Measuring blind spots in semantic segmentation. arXiv preprint
  arXiv:1904.03215  (2019)

\bibitem{kendall2017uncertainties}
Kendall, A., Gal, Y.: What uncertainties do we need in {Bayesian} deep learning
  for computer vision? In: Advances in neural information processing systems.
  pp. 5574--5584 (2017)

\bibitem{Weiss98}
Weiss, C.: Evaluation: Methods for Studying Programs and Policies. Prentice
  Hall (1998)

\bibitem{robotarium}
{Pickem}, D., {Glotfelter}, P., {Wang}, L., {Mote}, M., {Ames}, A., {Feron},
  E., {Egerstedt}, M.: The {Robotarium}: A remotely accessible swarm robotics
  research testbed. In: 2017 IEEE International Conference on Robotics and
  Automation (ICRA). pp. 1699--1706 (May 2017)

\bibitem{skinner}
Skinner, J., Garg, S., S{\"u}nderhauf, N., Corke, P., Upcroft, B., Milford, M.:
  High-fidelity simulation for evaluating robotic vision performance. In: 2016
  IEEE/RSJ International Conference on Intelligent Robots and Systems (IROS).
  pp. 2737--2744 (Oct 2016)

\bibitem{Montiel2015}
Mur-Artal, R., Montiel, J., Tardos, J.D.: Orb-slam: a versatile and accurate
  monocular slam system. Robotics, IEEE Transactions on  31(5),  1147--1163
  (2015)

\bibitem{lin2014microsoft}
Lin, T.Y., Maire, M., Belongie, S., Hays, J., Perona, P., Ramanan, D.,
  Doll{\'a}r, P., Zitnick, C.L.: Microsoft {COCO}: Common objects in context.
  In: European conference on computer vision. pp. 211--252740--755. Springer
  (2014)

\bibitem{ilsvrc:russakovsky2014imagenet}
Russakovsky, O., Deng, J., Su, H., Krause, J., Satheesh, S., Ma, S., Huang, Z.,
  Karpathy, A., Khosla, A., Bernstein, M., Berg, A.C., Fei-Fei, L.: {ImageNet
  Large Scale Visual Recognition Challenge}. International Journal of Computer
  Vision (IJCV)  115(3),  211--252 (2015)

\bibitem{sunderhauf2018limits}
S{\"u}nderhauf, N., Brock, O., Scheirer, W., Hadsell, R., Fox, D., Leitner, J.,
  Upcroft, B., Abbeel, P., Burgard, W., Milford, M., et~al.: The limits and
  potentials of deep learning for robotics. The International Journal of
  Robotics Research  37(4-5),  405--420 (2018)

\bibitem{thrun2005probabilistic}
Thrun, S., Burgard, W., Fox, D.: Probabilistic robotics. MIT Press, Cambridge,
  Mass. (2005)

\bibitem{wangaugpod}
Wang, C.W., Cheng, C.A., Cheng, C.J., Hu, H.N., Chu, H.K., Sun, M.: Augpod:
  Augmentation-oriented probabilistic object detection. CVPR workshop on the
  Robotic Vision Probabilistic Object Detection Challenge  (2019)

\bibitem{ammirato2019mask}
Ammirato, P., Berg, A.C.: A {Mask-RCNN} baseline for probabilistic object
  detection. arXiv preprint arXiv:1908.03621  (2019)

\bibitem{liteamgl}
Li, D., Xu, C., Liu, Y., Qin, Z.: {TeamGL} at {ACRV} robotic vision challenge
  1: Probabilistic object detection via staged non-suppression ensembling. CVPR
  workshop on the Robotic Vision Probabilistic Object Detection Challenge
  (2019)

\bibitem{MorrisonPDQ}
Morrison, D., Milan, A., Antonakos, A.: Uncertainty-aware instance segmentation
  using dropout sampling. CVPR workshop on the Robotic Vision Probabilistic
  Object Detection Challenge  (2019)

\end{thebibliography}
\bibliographystyle{splncs03_unsrt}
\end{document}